%% file: ms.tex
\documentclass[twocolumn,10pt]{asme2e}

\confshortname{IDETC/CIE 2022}
\conffullname{the ASME 2022 \\International Design Engineering Technical Conferences and\\
              Computers and Information in Engineering Conference}

\confdate{August 14-17}
\confyear{2022}
\confcity{St. Louis}
\confcountry{Missouri}

\papernum{DETC2022/MR-90595}

\usepackage[captionskip=2pt]{floatrow}
\newlength\Myfigwd

\usepackage{hyperref}
\usepackage{color}  %needed for color text

\newcommand{\realfield}[1]{\hbox{I \kern -.25em R}^{#1}}   %needed to
\newcommand {\mb}[1]{\mathbf{#1}}
\newcommand {\bs}[1]{\boldsymbol{#1}}
\newcommand{\uvec}[1]{\hat{\mathbf{#1}}}

\newcommand{\T}{^{\mathrm{T}}}  %shortcut for transpose
\newcommand{\rmd}{\textrm{d}}  %shortcut for derivative

\usepackage[%
font=small,
labelfont=bf,
singlelinecheck=off]{caption}
\usepackage{array}
\newcolumntype{P}[1]{>{\centering\arraybackslash}p{#1}}
\newcolumntype{M}[1]{>{\centering\arraybackslash}m{#1}}

\newcommand{\myhigh}[1]{\textit{\ul{#1}}}

\usepackage{lipsum}
\usepackage{graphicx}
\usepackage{comment}
\usepackage{gensymb}
\usepackage{amsmath}
\usepackage{amssymb}
\usepackage{mathtools}
\usepackage{commath}
\usepackage{etoolbox}
\usepackage{caption}
\usepackage{subcaption}
\usepackage{siunitx}
\usepackage[compress]{cite}
\usepackage{subcaption}
\renewcommand{\citedash}{--}
\usepackage{float}
\floatstyle{plaintop}
\restylefloat{table}
\usepackage{enumitem}
\usepackage{arydshln}
\usepackage{soul}
\usepackage{longtable}
\graphicspath{
	{figures/png/}
	{figures/pdf_fig/}}

\begin{document}
\input{content/title-authors}
\maketitle    

\input{content/abstract}
\input{content/asme_nomenclature}

\input{content/intro}
\input{content/mechanism}
\input{content/kinematics}

\input{content/experiments}
\input{content/discussion}
\input{content/conclusion}

\bibliographystyle{asmems4}
\bibliography{%
	bib/continuum_robot_general,%
	bib/aerial_manipulation,%
	bib/aerial_manipulation_compliant,%
	bib/justin_paper,%
	bib/long_added_references,%
	bib/kinematics}

\end{document}

%% file: content/title-authors.tex
\author{Qianwen Zhao\\[2pt]
	{\tensfb Guoqing Zhang}\\[2pt]
	{\tensfb Hamidreza Jafarnejadsani}\\[2pt]
	{\tensfb Long Wang}\thanks{Address all correspondence for other issues to this author.}\\[2pt]
	\affiliation{Department of Mechanical Engineering\\
		Schaefer School of Engineering \& Science\\
		Stevens Institute of Technology \\
		Hoboken, NJ, 07030 USA\\
		qzhao10@stevens.edu, gzhang21@stevens.edu, hjafarne@stevens.edu, lwang4@stevens.edu
	}
}

%%% You need to remove 'DRAFT: ' in the title for the final submitted version.
\title{A Modular Continuum Manipulator for Aerial Manipulation and Perching}

%% file: content/abstract.tex
\begin{abstract}
{\it Most aerial manipulators use serial rigid-link designs, which results in large forces when initiating contacts during manipulation and could cause flight stability difficulty. This limitation could potentially be improved by the compliance of continuum manipulators.\par
To achieve this goal, we present the novel design of a compact, lightweight, and modular cable-driven continuum manipulator for aerial drones. We then derive a complete modeling framework for its kinematics, statics, and stiffness (compliance). The framework is essential for integrating the manipulator to aerial drones. Finally, we report preliminary experimental validations of the hardware prototype, providing insights on its manipulation feasibility. Future work includes the integration and test of the proposed continuum manipulator with aerial drones.
}
\end{abstract}

%% file: content/asme_nomenclature.tex
\begin{nomenclature}
	\small
	\entry{Frame \{B\}}{The base frame with $\mb{b}$ located at the center of the base, $\uvec{x}_b$ passing through the first wire and $\uvec{z}_b$ perpendicular to the base.}
	\entry{Frame \{1\}}{Characterizes the plane in which the continuum segment bends and it is obtained by a rotation of $\delta$ about $\uvec{z}_b$. Unit vector $\uvec{x}_1$ is along the projection of the central axis on the plane of the base and $\uvec{z}_1=\uvec{z}_b$.}
	\entry{Frame \{E\}}{It is defined with $\uvec{z}_e$ as the normal to the end surface and $\uvec{x}_e$ is the intersection of the bending plane and the end disk top surface.}
	\entry{Frame \{G\}}{Frame \{G\} defined with $\uvec{z}_g=\uvec{z}_e$ and its $\uvec{x}_g$ passing through the $1^\text{st}$ wire. It can be obtained by a rotation angle $\delta_e$ about $\uvec{z}_e$ which is the unit vector normal to the end disk. This angle is given by $\delta_e = -\delta$.}
	\entry{$i$}{Index of the tendons, $i$=1,2,3,4.}
	\entry{$r$}{Radius of the pitch circle where the tendons are distributed.}
	\entry{$\delta$}{Bending plane angle.}
	\entry{$\theta(s)$}{Bending angle; the angle of the $\uvec{z}_1$ or $\uvec{z}_b$ to the tangent at the s arc length position in the bending plane. $\theta(L)=\theta$.}
	\entry{$L,L_{i}$}{Length of central backbone, Length of $i^{th}$ tendon measured from the base disk to the end disk.}
	\entry{$q_{i}$}{The displacement of $i^{th}$ tendon.}
	\entry{$\beta$}{Division angle of the tendons along the circumference of the pitch circle,$\beta$=$\pi$/2.}
	\entry{${ }^{a} \mathbf{T}_{b}$}{Homogeneous transformation matrix describing orientation and position of frame\{b\} with respect to frame\{a\}.}
	\entry{${ }^{a} \mathbf{p}_{b},^{a} \mathbf{R}_{b}$}{Position vector and rotation matrix of frame\{b\} with respect to frame\{a\}, respectively.}
	\entry{$\psi,\dot{\psi}$}{Configuration space vector and corresponding time derivative,$\psi = [\theta, \delta]^{\mathrm{T}}$.}
	\entry{$\mb{q},\dot{\mb{q}}$}{Tendon displacement vector and corresponding time derivative,$\mathbf{q} = [ q_1, q_2, q_3, q_4]^{\mathrm{T}}$.}
	\entry{$\dot{\mb{x}}$}{ vector in task space.}
	\entry{$\mb{J}_{\mathbf{q}\psi}$}{A Jacobian matrix mapping the configuration space velocities to joint velocities.}
	\entry{$\mb{J}_{\mathbf{\mb{x}}\psi}$}{A Jacobian matrix mapping the configuration space velocities to task space velocities.}
	\entry{$\mb{J}_{\mathbf{v}\psi}$}{A Jacobian matrix mapping the configuration space velocities to linear velocities.}
	\entry{$\mb{J}_{\mathbf{\omega}\psi}$}{A Jacobian matrix mapping the configuration space velocities to angular velocities.}
	\entry{$\mb{E},\nabla \mb{E}$}{Potential elastic energy of the continuum arm and corresponding gradient with respect to the configuration space}
	\entry{$\bs{\tau}$}{Tendon actuation force,$\bs{\tau}=\left[\tau_{1}, \ldots, \tau_{4}\right]^{\mathrm{T}}$.}
	\entry{$E_{p}, E_{T}$}{Young's modulus of the central NiTi backbone and Young's modulus of the tendon, respectively.}
	\entry{$I_{p}$}{Cross-sectional second moment of inertia of the NiTi backbone.}
	\entry{$A$}{Cross-sectional area of the tendon.}
	\entry{$\mb{w}_{\text{ext}},\mb{F}_{\text{ext}}$}{External wrench and force applied to the end disk tip of the continuum arm, respectively.}
	\entry{$\mb{K}_X,\mb{K}_{\psi}$}{Task space stiffness and configuration space stiffness, respectively.}
	\entry{$\mb{K}_q$}{Stiffness matrix of tendons}
	\entry{$\mb{F}^{*}$}{Generalized force applied to continuum arm in configuration space}
\end{nomenclature}

%% file: content/intro.tex
\section{Introduction}
\label{section:intro}
According to the Federal Aviation Administration (FAA) aerospace forecast release \cite{FAA_Forecast}, there were approximately 1.1 million unmanned aerial vehicles (UAVs), or so-called drones, in the United States as of December 31, 2017. Today, multirotor UAVs are household products and are widely used for photography, mapping and inspection, or simply for recreation. Despite their success in perceptual tasks, the mainstream uses of UAVs do not involve physical interactions with objects or the environment. 
The ability to physically manipulate objects and environments unlocks new tasks that aerial drones could be capable of. To this end, many researchers have spent great efforts to enhance aerial drones with manipulation capabilities \cite{Pounds2011,Mellinger2011,Korpela2013,Scholten2013,Korpela2014,Garimella2015,Anzai2018,Nguyen2018,Ruggiero2018,Baraban2020,Kim2020}, and the investigations span a variety of topics such as aerial grasping, aerial interaction force, flight and payload dynamics, cooperative control, model-predictive control, and haptics-based manipulation. \par
Compliance has been introduced to manipulation tasks as an advantage because it can potentially provide robustness to errors. In the case of aerial manipulation, compliance can also help increase the safety margin by softening the collision impact when initiating contacts during manipulation. Recent works have explored the use of compliance in aerial manipulation using soft manipulators, compliant manipulators, and underactuated hands \cite{Pounds2011,Yuksel2015,Suarez2015,Suarez2018,Samadikhoshkho2020,Fishman2021,Fishman2021a}. \par
Continuum manipulators are an unconventional class of robots having inherited compliance in their bio-inspired structures such as an elephant's trunk or an octopus's tentacles \cite{Hannan2003,Liu2019,Yeshmukhametov2019,Zheng2012,Zheng2013,Cianchetti2012,Cianchetti2015,Gravagne2002,Gravagne2002a,Simaan2004,Conrad2015,Conrad2017,simaan2018medical,McMahan2006,Rone2018,Rone2019,sitler2022modular}. In this work, we propose to introduce continuum manipulators to aerial manipulation, taking advantage of the mechanism compliance, and the compact and lightweight design.

\subsection{Related Work}
\noindent \myhigh{Aerial Manipulation.}\quad  Pounds \textit{et~al}. \cite{Pounds2011} and Mellinger \textit{et~al}. \cite{Mellinger2011} investigated aerial grasping control methods considering payload dynamics and demonstrated on a helicopter platform and on a quadrotors, respectively.
Scholten \textit{et~al}. presented interaction force control of an aerial drone equipped with a Delta parallel manipulator\cite{Scholten2013}. 
Korpela \textit{et~al}. investigated dynamic stability of a mobile UAV equipped with two 4-DoF manipulators in \cite{Korpela2013} and continued to develop a framework for the valve turning task in \cite{Korpela2014}. 
Nguyen \textit{et~al}. presented a novel method for aerial manipulation where multiple quadrotors are connected via passive spherical joints and generate thrusts for manipulation tasks collectively \cite{Nguyen2018}.
Anzai \textit{et~al}. developed a transformable multirotor robot with closed-loop multilinks structure that can adapt to shapes of large objects to be grasped \cite{Anzai2018}.
Garimella and Kobilarov introduced model-predictive control in aerial pick-and-place tasks \cite{Garimella2015}, and Baraban \textit{et~al}. continued to explore an adaptive parameter estimation method for unknown mass of objects \cite{Baraban2020}.
Kim and Oh developed a haptics-based testing-and-evaluation platform for drone-based aerial manipulation \cite{Kim2020}.\par

\noindent\myhigh{Aerial Manipulation with Compliance.}\quad A compliant underactuated manipulator was used in Pounds \textit{et~al}.'s work \cite{Pounds2011}. 
Suarez \textit{et~al.} developed a lightweight compliant arm for aerial drones and proposed its usage in payload mass estimation\cite{Suarez2015}. The same authors extended the compliant arm design for dual-arm aerial manipulation \cite{Suarez2018}. 
Y\"uksel \textit{et~al.} presented another lightweight compliant arm design using a cable-driven flexible-joint for aerial manipulation \cite{Yuksel2015}. 
Samadikhoshkho \textit{et~al.} introduced a modeling framework for considering the use of continuum manipulators for aerial manipulation tasks and reported findings in a simulated environment \cite{Samadikhoshkho2020}.
Fishman \textit{et~al.} prototyped a soft gripper for aerial manipulation, developed algorithms for trajectory optimization considering the soft gripper modeling, and demonstrated aerial grasping in both simulations and physical experiments \cite{Fishman2021,Fishman2021a}.
\par

\noindent \myhigh{Continuum Manipulators.}\quad Continuum manipulators have been widely used in biomedical applications \cite{Simaan2004, Conrad2015, Conrad2017, simaan2018medical} because of their flexibility and compliance, enabling safe interaction with the anatomy. They are also good candidates for field robotics applications.
McMahan \textit{et~al.} demonstrated the field capabilities and manipulation of a pneumatic continuum manipulator attached to a mobile robot platform in \cite{McMahan2006}. 
Rone \textit{et~al.} used a continuum tail to stabilize a walking mobile robot \cite{Rone2018,Rone2019}.
Sitler and Wang recently presented a modular continuum manipulator design for free-floating underwater manipulation tasks \cite{sitler2022modular}.

\begin{figure}
	\centering
	\includegraphics[width=0.8\columnwidth]{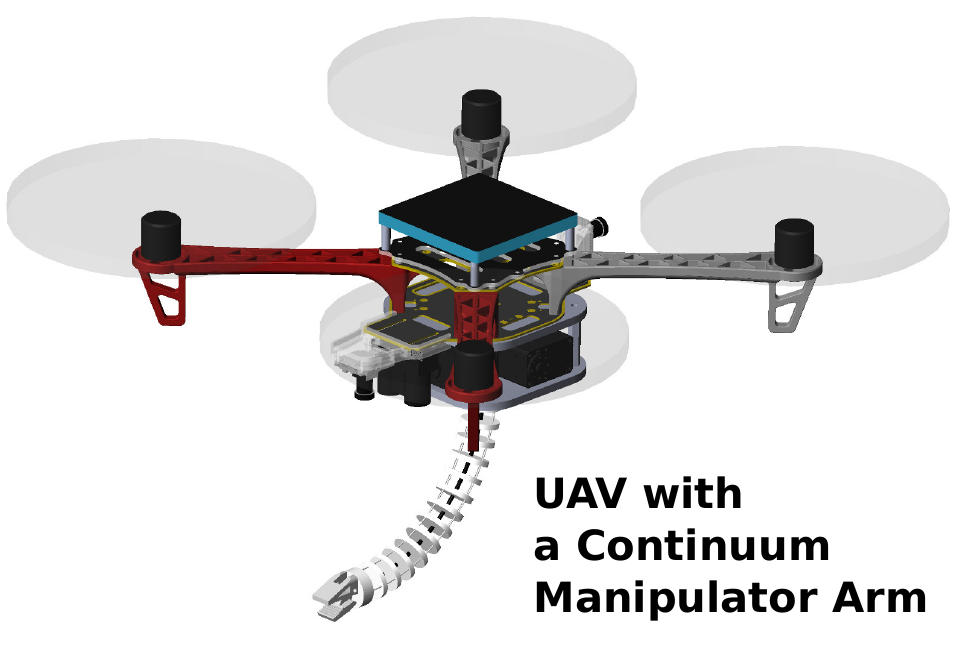}
	\caption{Schematic of an unmanned aerial vehicle(UAV) integrated with a continuum manipulator arm}
	\label{fig:schematic of overall system}
\end{figure}

\subsection{Contribution of this work}
Aiming at introducing continuum manipulators into aerial manipulation (illustrated in Fig.~\ref{fig:schematic of overall system}), the contributions of this work include the following:
\begin{enumerate}
	\item We present a novel design of a compact, lightweight, and modular cable-driven continuum manipulator for aerial drones.
	\item We derive a complete modeling framework for kinematics, statics, and stiffness (compliance), which is essential for integrating to aerial drones.
	\item We report preliminary experimental validations of the hardware prototype, providing insights on its integration feasibility.
\end{enumerate}

%% file: content/mechanism.tex
\section{Mechanism Description}
The mechanical structure of the proposed modular continuum robot comprises of the arm section and the driving mechanism housing section. We start with overall design considerations for aerial applications, followed by design details of the continuum arm and the actuation unit.

\subsection{Overall Design Considerations}
One advantage of a continuum arm, compared to industrial robots, is its mechanical compliance. Safe robot-human and robot-environment interactions can benefit from the robot being able to adapt its shape. 
Unlike conventional robot arms with revolute and prismatic joints chained together, the robot structure and driving mechanism of continuum robots have a wide range of variability. 
The mechanical structure of the continuum robot can be categorized into the soft structure continuum arm \cite{JONES2004,Fras2018}, and multi-disks with single/multi-segment \cite{Xu2021,Neumann2016}, and the most commonly seen driving mechanisms are either motor-actuated tendon-driven robot \cite{Xu2021,Renda_2012,Neumann2016} or soft structure pressurized by pneumatic actuators \cite{JONES2004,McMahan_2005}. 
With advantages such as simplified kinematic computations and high accuracy in determining the end-effector position, a tendon-driven continuum robot (TDCR) can perform aerial manipulation and perching applications. 
We use disks to keep multiple tendons parallel, which improves its mechanical compliance.

Another design consideration is lightweight due to UAV's payload limitation. The use of relatively small and compact servo motors as actuators will be discussed further in later paragraphs in this section. To achieve lightweight and fast prototyping, most components of the arm and actuation unit housing are 3D printed with Polylactic Acid (PLA). The lightweight compliant structure is stiffened by using NiTi cables as continuum backbones. 

Such a manipulator design can be adapted to multiple drone platforms and to various tasks. 
The aerial manipulator system consists of a continuum arm, a driving mechanism and an actuation unit housing, and its adapter interface to the UAV platform.
Thanks to the stacking up arm architecture, the continuum arm range of motion can be extended by adding more spacer disks and longer cables correspondingly before the end disk, without effecting the driving mechanism and basic control logic.
Moreover, the modular design allows for adding additional segments to the end disk of the previous one. This enables the arm to achieve more articulating poses such as an "s" shape in multiple bending planes in the future. \par
\begin{figure}[!h]
    \centering
    \includegraphics[width=0.97\columnwidth]{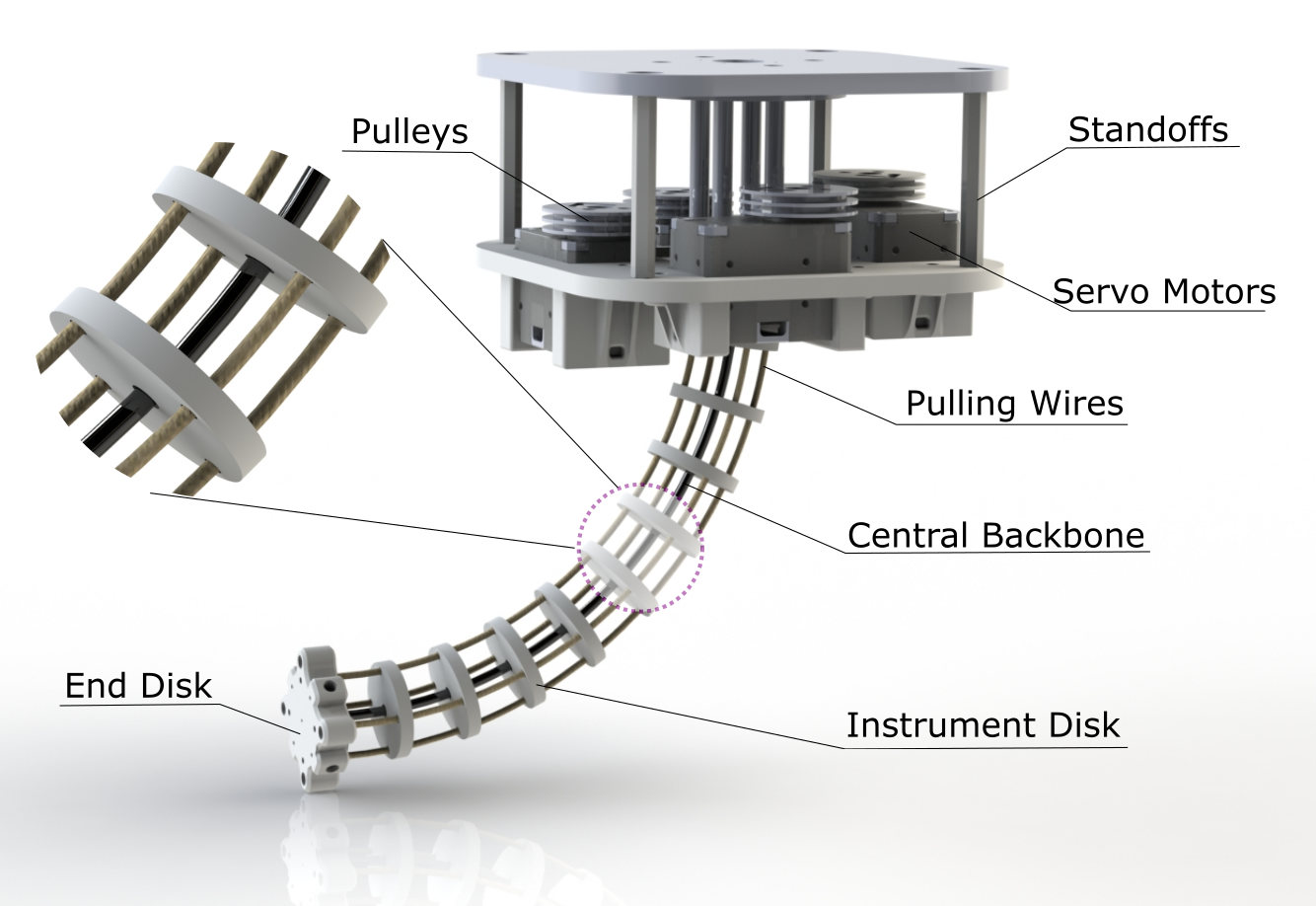}
    \caption{Full Assembly of the proposed modular continuum arm and a close-up view: primary backbone is in black and four tendons are in beige.}
    % The primary backbone is in black; four pulling tendons are in beige; and all end disk, base disk (concealed under the electronics housing), instrument disks, and electronics housing are 3D printed in white PLA.
    \label{fig:fullassembly}
\end{figure}

\subsection{Continuum Arm Design}
Figure~\ref{fig:fullassembly} depicts the proposed modular continuum arm for aerial manipulation. The arm consists of a central backbone, four tendons, nine circular structural disks, an end disk, and a base disk integrated with cable routing mechanism. 
Steel reinforced epoxy is applied to secure the central backbone to keep it at a constant length while bending. 
Four tendons are rigidly connected at the end disk. Distances between adjacent disks are designed to be constants that allows the arm articulate. Fishing wires are used as tendons due to their durability. 
A closer inspection of the arm can be found in the top left corner of Fig.~\ref{fig:fullassembly}.
To actuate the motion of the continuum manipulator, four Dynamixel XL430-W250-T servo motors with 57.2g each are chosen to pull four fishing wires. The arm prototype being built is 420 grams in total, including both arm and actuation unit. It is feasible for a relatively lightweight commercial drones to carry. \par

\subsection{Actuation Unit}
A 12V custom motor shield is installed inside the continuum arm housing. The motor shield uses Teensy~4.0 as the main micro-controller due to its fast processing speed and compact form. By daisy-chaining four motors together, only one cable connection to the motor shield is necessary. Each fishing wire passes from the arm to the actuation unit through a bearing and wrapped on a motor pulley. During arm articulation, certain motor(s) draw wires while their opposites release wires, and such coordination drives the end-effector to its designated pose. Detailed motor coordination and kinematics modeling will be discussed in the next section.

%% file: content/kinematics.tex
\section{Kinematics, Statics and Stiffness Modeling}
The nomenclature described earlier is referenced to facilitate the modeling of kinematics, mechanics and stiffness of the continuum structure.  \par
\subsection{Kinematics}
\begin{figure}
    \centering
    \includegraphics[width=0.97\columnwidth]{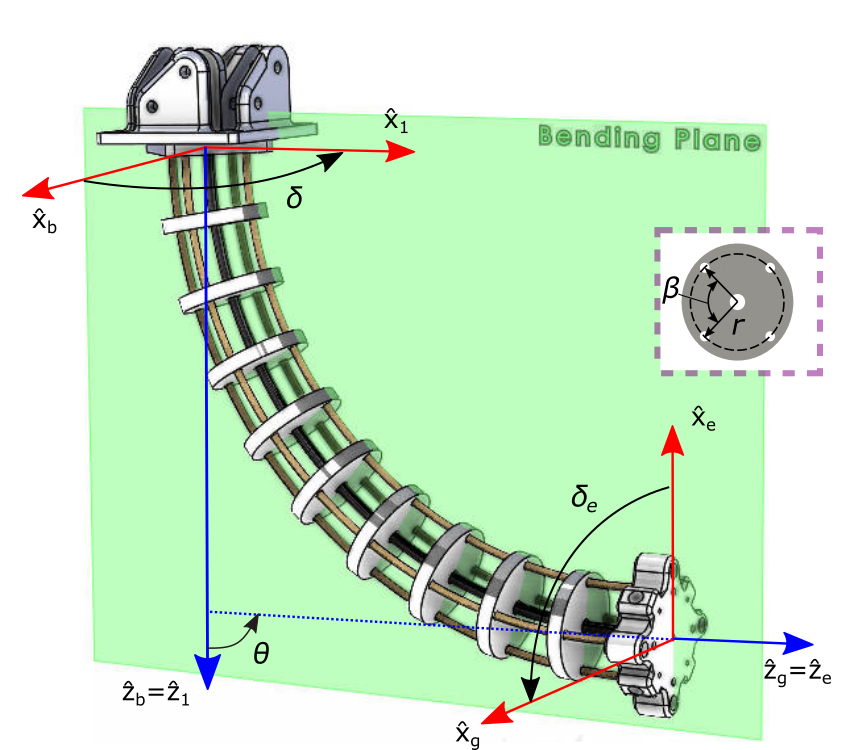}
    \caption{Schematic of the proposed modular continuum arm.}
    \label{fig:schematic of kinematics}
\end{figure}

Kinematic modeling of a robot arm usually starts with identifying relevant kinematic spaces. A continuum arm kinematics uses three kinematic spaces: the configuration space, the joint space, and the task space. 
The configuration space is defined as a combination of the in-plane bending angle $\theta$ and the bending plane angle $\delta$, with their vector form as $\psi = [\theta, \delta]^{\mathrm{T}}$. The configuration space is illustrated in Fig.~\ref{fig:schematic of kinematics}. The home (straight) configuration of the proposed continuum arm is \(\psi = [0^\circ, 0^\circ]^{\mathrm{T}}\). 
Motor positions are denoted by \(q_i\) and they form the joint space. In our case, four motors are used to pull four tendons, \(\mathbf{q} = [ q_1, q_2, q_3, q_4]^{\mathrm{T}} \).
Task space indicates the position and orientation of the manipulator's end-effector with respect to the base coordinate. The corresponding homogeneous transformation matrix is as follows:
\begin{equation}
{ }^{b} \mathbf{T}_{g}={ }^{b} \mathbf{T}_{1}{ }^{1} \mathbf{T}_{e}{ }^{e} \mathbf{T}_{g}
\end{equation}
where\par
\begin{equation}
\begin{aligned}
&{ }^{b} \mathbf{T}_{1}=\left[\begin{array}{cc}
\operatorname{RotZ}(\delta) & \mathbf{0} \\
\mathbf{0} & 1
\end{array}\right] \\
&{ }^{1} \mathbf{T}_{e}=\left[\begin{array}{cc}
\operatorname{RotY}(\theta) & { }^{1} \mathbf{p}_{e} \\
\mathbf{0} & 1
\end{array}\right], \quad
{ }^{1} \mathbf{p}_{e}=\frac{L}{\theta}\left[\begin{array}{c}
1-\cos \theta \\[2pt]
0 \label{eqn:kin_p_e}\\[2pt]
\sin \theta
\end{array}\right] \\
&{ }^{e} \mathbf{T}_{g}=\left[\begin{array}{cc}
\operatorname{RotZ}(-\delta) & \mathbf{0} \\
\mathbf{0} & 1
\end{array}\right]
\end{aligned}
\end{equation}
Mappings among these three spaces allows for arm motion control motion. \par
\subsubsection{Configuration to Joint Space}
Inverse kinematics relating the configuration space and joint space is shown as following: 
% you may use \; or \, to add space between two symbols
\begin{equation}
    \begin{aligned}
        q_1 &= r\,\cos(\delta)\,\theta \\
        q_2 &= r\,\cos(\delta+\beta)\,\theta \\
        q_3 &= r\,\cos(\delta+2\beta)\,\theta \\
        q_4 &= r\,\cos(\delta+3\beta)\,\theta
    \end{aligned}
\end{equation}
where $ r $ represents the radius of the pitch circle along which the four tendons are circumferentially distributed.

Taking the derivative of (3), the instantaneous inverse kinematics is obtained:
\begin{equation}
    \mathbf{\dot q} = \mb{J}_{\mathbf{q}\psi} \dot{\psi}
\end{equation}
where
\begin{equation}
\mathbf{J}_{\mathbf{q} \psi} =
\begin{bmatrix}
r\, \cos(\delta)                   & -r\, \sin(\delta)\theta \\[6pt]
r\, \cos(\delta+\beta)    & -r\, \sin(\delta+\beta)\theta \\[6pt]
r\, \cos(\delta+2\beta)               & -r\, \sin(\delta+2\beta)\theta \\[6pt]
r\, \cos(\delta+3\beta)    & -r\, \sin(\delta+3\beta)\theta
\end{bmatrix}
\end{equation}
\subsubsection{Configuration to Task Space}

Similarly, the instantaneous direct kinematics can be expressed as
\begin{equation}
\dot{\mb{x}}=\mb{J}_{\mb{x} \psi} \dot{\boldsymbol{\psi}}, \qquad
\dot{\mb{x}}\triangleq\left[\mb{v}\T,\bs{\omega}\T\right]\T,\quad
\mb{J}_{\mb{x} \psi}\in \realfield{6 \times2}
\end{equation}
where $\dot{\mb{x}}$ represents a twist that consists of a linear velocity $ \mb{v} $ and an angular velocity $ \bs{\omega} $, and 
$\mathbf{J}_{\mathbf{x} \psi}= \begin{bmatrix}
\mathbf{J}_{\mathbf{v} \psi}^{\mathrm{T}} &\quad
\mathbf{J}_{\mathbf{\omega} \psi}^{\mathrm{T}}
\end{bmatrix}^{\mathrm{T}}$ is the geometric Jacobian matrix that consists of a linear velocity partition and an angular velocity partition. Matrix $\mathbf{J}_{\mathbf{v} \psi}$ can be obtained by differentiating the position ${ }^{b} \mathbf{p}_{e}$. \par 
%怕人
Based on (\ref{eqn:kin_p_e}), ${ }^{b} \mathbf{p}_{e}$ may be expressed as:
\begin{equation}
{ }^{b} \mathbf{p}_{e}=
\operatorname{RotZ(\delta)}{ }^{1} \mathbf{p}_{e} = \frac{L}{\theta}\left[\begin{array}{c}
\cos(\delta)\,\left(1-\cos\theta\right) \\[2pt]
\sin(\delta)\,\left(1-\cos\theta\right) \\[2pt]
\sin(\theta)
\end{array}\right]
\end{equation}
Thereby, $\mathbf{J}_{\mathbf{v} \psi}$ is given by
\begin{equation}
\mathbf{J}_{\mathbf{v} \psi}=L\left[\begin{array}{cc}
\cos(\delta)\, \frac{\theta \sin(\theta)+\cos(\theta)-1}{\theta^{2}} & -\frac{\sin(\delta)\,\left(1-\cos(\theta)\right)}{\theta} \\[6pt]
\sin(\delta)\, \frac{\theta \sin(\theta)+\cos(\theta)-1}{\theta^{2}} & \frac{\cos(\delta)\,\left(1-\cos(\theta)\right)}{\theta} \\[6pt]
\frac{\theta \cos(\theta)-\sin(\theta)}{\theta^{2}} & 0
\end{array}\right]
\end{equation}
\par Matrix $\mathbf{J}_{\mathbf{\omega} \psi}$ can be obtained by directly formulating the angular velocity in base coordinate shown as follows
\begin{equation}
{ }^{b} \boldsymbol{\omega}_{g}=\dot{\delta}^{b} \hat{\mathbf{z}}_{b}+{ }^{b} \mathbf{R}_{1}\left(\dot{\theta}^{1} \hat{\mathbf{y}}_{1}+{ }^{1} \mathbf{R}_{e}\left(-\dot{\delta}^{e} \hat{\mathbf{z}}_{e}\right)\right) = \mathbf{J}_{\omega \psi} \dot{\boldsymbol{\psi}}
\label{eqn:J_omega_psi_derivation}
\end{equation}
where $^{b}\hat{\mathbf{z}}_{b} = [0, 0, 1]^{\mathrm{T}}$, $^{1}\hat{\mathbf{y}}_{1} = [0, 1, 0]^{\mathrm{T}}$, $^{e}\hat{\mathbf{z}}_{e} = [0, 0, 1]^{\mathrm{T}}$ in their respective coordinates. ${ }^{b}\mathbf{R}_{1}$ and ${ }^{1} \mathbf{R}_{e}$ can be obtained from (\ref{eqn:kin_p_e}).
By rearranging (\ref{eqn:J_omega_psi_derivation}) in a vector format, matrix $\mathbf{J}_{\omega \psi}$ can be given by
\begin{equation}
\mathbf{J}_{\omega \psi}=\left[\begin{array}{cc}
-\sin(\delta)& -\cos(\delta)\,\sin(\theta)   \\
\cos(\delta) & -\sin(\delta)\,\sin(\theta)  \\
0& 1-\cos(\theta) 
\end{array}\right]
\end{equation}
\subsection{Statics}
We use the virtual work analysis to derive the statics in this section. We take the potential elasticity energy into account. When the continuum arm bends, the central NiTi backbone stores the potential elastic energy, given by:

\begin{equation}
\mb{E}=\int_{L} \frac{E I}{2}\left(\frac{d \theta}{d s}\right)^{2} d s=\theta^{2}\frac{E_{p} I_{p}}{2 L}
\end{equation}
\par Following the derivation method in \cite{simaan2005snake,Goldman2014}, the statics model of the continuum arm can be described as:
\begin{equation}
\mb{J}_{\mb{q} \psi}^{\mathrm{T}} \bs{\tau}+\mb{J}_{\mb{x} \psi}^{\mathrm{T}} \mathbf{w}_{\text{ext}}=\nabla \mb{E} 
\end{equation}
where $\bs{\tau}$ represent the actuation forces for the tendon pulling, $\mathbf{w}_\text{ext}$ is the 6-dimensional external wrench applied to the end-disk of the continuum arm, $\nabla \mb{E}$ describes the gradient of the elastic energy with respect to the configuration space change.\par 
By differentiating (11), $\nabla \mb{E}$ takes the form:
\begin{equation}
\begin{aligned}
&\nabla \mb{E}
&=\left[\begin{array}{c}
\frac{\theta E_{p} I_{p}}{L}\\
0
\end{array}\right]
\end{aligned}
\end{equation}
\input{content/stiffness_modeling}

%% file: content/stiffness_modeling.tex
\subsection{Stiffness}
Stiffness modeling of a continuum manipulator is essential for the case of using it in aerial manipulation. Without considering the reaction moment and orientational displacement, we limit the scope of the stiffness modeling problem to incremental force and positional displacement. \par
\begin{figure}[!h]
	\centering
	\includegraphics[width=0.75\columnwidth]{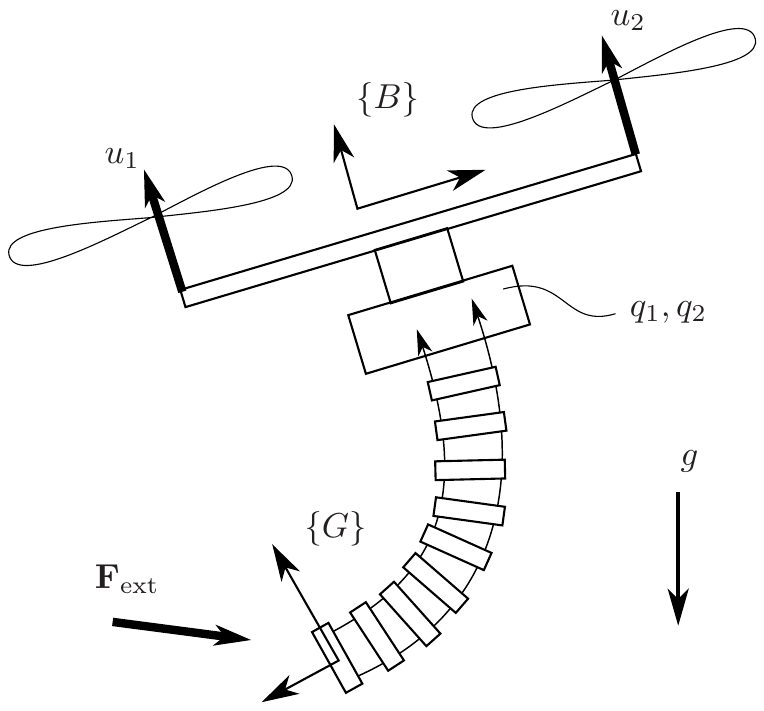}
	\caption{Stiffness modeling of a continuum manipulator for a drone}
	\label{fig:stiffness_modeling_aerial}
\end{figure}\par
As illustrated in Fig.~\ref{fig:stiffness_modeling_aerial}, let $ \mb{F}_\text{ext} $ and $ {}^b\mb{p}_g $ both be $ (3\times1) $ vectors, representing an external force and the gripper position expressed in the manipulator base frame, respectively. Thereby, the ultimate goal of stiffness modeling is to capture a $ 2^{\text{nd}} $-order gradient tensor, $ \mb{K}_X $, defined as:
\begin{equation}
	\mb{K}_X \triangleq \dfrac{\partial  (\mb{F}_{\text{ext}})}{\partial  ({}^b\mb{p}_g)}, \qquad \mb{K}_X\in\realfield{3\times3}
\end{equation}\par
The stiffness matrix $ \mb{K}_X $ (the $ 2^{\text{nd}} $-order gradient tensor) defines the linearized local relations between small force change $ \rmd\mb{F} $ and small deformation $ \rmd \mb{p} $. However, the derivation of $ \mb{K}_X $ is nontrivial for continuum robots due to the highly nonlinear mappings among the three kinematic spaces as discussed earlier. \par
A stiffness matrix in the configuration space can be found only assuming that the external force is very small \cite{Goldman2014}, and it is thereby defined as:
\begin{equation}
	\mb{K}_\psi \triangleq \frac{\partial \mb{F}^{*}}{\partial \bs{\psi}}, \qquad \mb{K}_\psi\in\realfield{2\times2}
\end{equation}
where $ \mb{F}^{*} $ denotes the generalized force in configuration space when the external wrench (including force) is very small. By considering the static equilibrium state from (12), we can express such a generalized force applied to the continuum arm in configuration space, as:
\begin{equation}
 \mb{F}^{*} =\nabla \mb{E}-\mb{J}_{\mb{q} \psi}^{\mathrm{T}} \bs{\tau}=\mb{J}_{\mb{x} \psi}^{\mathrm{T}} \mathbf{w}_{\text{ext}}\approx\mb{0},\qquad 
 \mb{F}^{*} \in \realfield{2\times1}
\end{equation}\par 
Following \cite{Goldman2014}, one could find the expression of the aforementioned configuration space stiffness matrix, $ \mb{K}_\psi $, as:
\begin{equation}
\mathbf{K}_{\boldsymbol{\psi}}=\mathbf{H}_{\boldsymbol{\psi}}-\left[\frac{\partial}{\partial \boldsymbol{\psi}}\left(\mathbf{J}_{\mathbf{q} \psi}\right)\T\right] \boldsymbol{\tau}-\mathbf{J}_{\mathbf{q} \psi}^{\mathrm{T}} \mathbf{K}_{\mathbf{q}} \mathbf{J}_{\mathbf{q} \psi}
\end{equation}
where 
\begin{align}
	& \mb{H}_{\bs{\psi}} = \left[\begin{array}{cc}
\dfrac{\partial^{2} \mb{E}}{\partial \theta^{2}} &
\dfrac{\partial^{2} \mb{E}}{\partial \theta \partial \delta}\\[8pt]
\dfrac{\partial^{2} \mb{E}}{\partial \delta \partial \theta}&
\dfrac{\partial^{2} \mb{E}}{\partial \delta^{2}}
\end{array}\right]= 
\left[\begin{array}{cc}
\dfrac{ E_{p} I_{p}}{L} &0\\[8pt]
0& 0
\end{array}\right]
\\
& \mb{K}_q= \operatorname{diag}^{4}\left(\left[\frac{E_{T} A}{L}, \ldots, \frac{E_{T} A}{L}\right]\right)
\end{align}

Note that in the above equation, $\left[\frac{\partial}{\partial \boldsymbol{\psi}}\left(\mathbf{J}_{\mathbf{q} \psi}\right)\T\right]$ is a $3^{rd}$-order tensor (or a ``3D matrix''), and the multiplication is facilitated as:
\begin{equation}
    \left[\dfrac{\partial}{\partial \boldsymbol{\psi}}\left(\mathbf{J}_{\mathbf{q} \psi}\right)\T\right] \boldsymbol{\tau}\quad =\quad
    \left[
        \begin{array}{c;{2pt/2pt}c}
	         \dfrac{\partial \mb{J}_{\mb{q} \psi}^{\mathrm{T}}}{\partial {\theta}} \; \bs{\tau} \;\;  & \;\;  
	         \dfrac{\partial \mb{J}_{\mb{q} \psi}^{\mathrm{T}}}{\partial {\delta}} \; \bs{\tau} \;\;
	        \end{array}
    \right]
\end{equation}

We can find the expression of $ \mb{K}_X $ if we keep using the small force assumption. Let us express the small force using the generalized force, as:

\begin{equation}
 	\mb{F}^{*}=\mathbf{J}_{\mathbf{v} \boldsymbol{\psi}}^{\mathrm{T}} \mb{F}_{\text{ext}} \quad \rightarrow \quad \mb{F}_\text{ext}=\left(
	\mb{J}_{\mb{v} \psi}^{\mathrm{T}}
	\right)^{\dagger}
	\mb{F}^{*}
\end{equation}

\begin{figure*}[!h]
\setlength\Myfigwd{0.75\columnwidth}
\floatbox[{\capbeside\thisfloatsetup{
capbesideposition={right,center},
capbesidewidth=\dimexpr\linewidth-\Myfigwd-3em\relax}}]{figure}[\FBwidth]
{\caption{Stiffness test results:(a) Inward radial loading; (b) Outward radial loading.}
\label{fig:stiffness results}}
{\includegraphics[width=\Myfigwd]{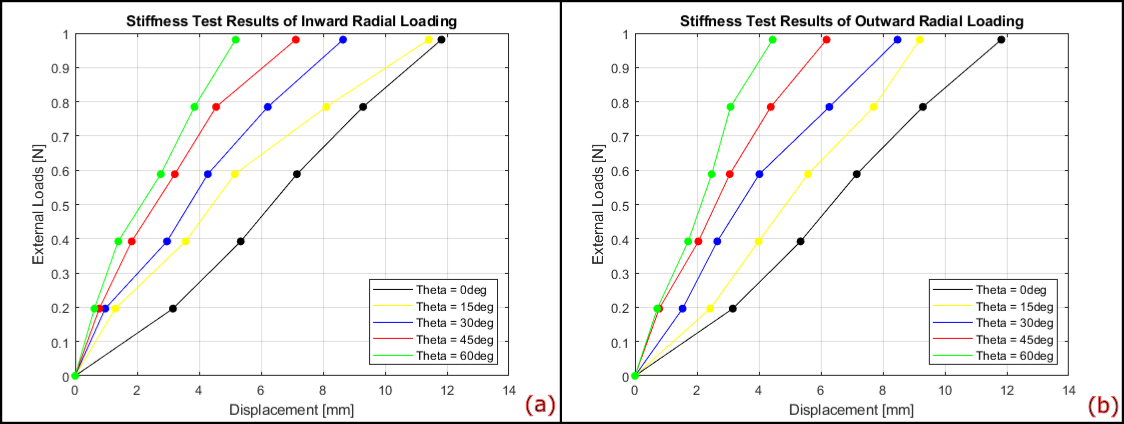}}
\vspace{-6mm}
\end{figure*}
\begin{figure}
    \centering
    \includegraphics[width=1\textwidth]{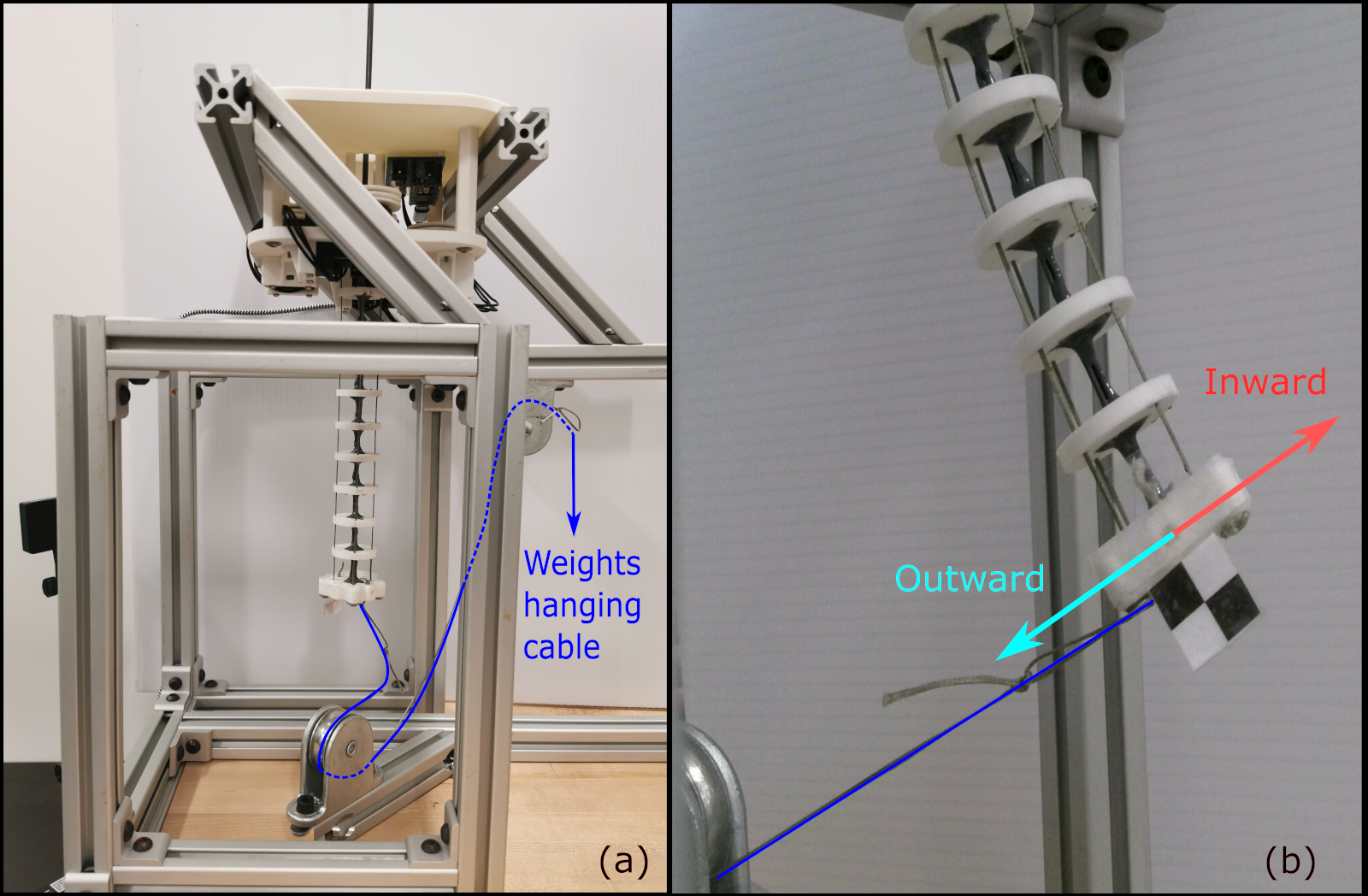}
    \caption{(a) Test setup of aerial continuum arm stiffness evaluation under a known load. The top plate of the arm system is fixed with the testing structure, and a known load is applied at the tip of the arm. (b) Indication of inward and outward radial loading.}
    \label{fig:stiffness test setup}
\end{figure}

\par Thus, the stiffness matrix under task space is given by
\begin{equation}
	\begin{array}{rl}
		\mb{K}_{X} & =\frac{\partial \mb{F}_{\text{ext}}}{\partial ({}^b\mb{p}_g)}
		= \frac{\partial \left(
			\left(
			\mb{J}_{\mb{v} \psi}^{\mathrm{T}}
			\right)^{\dagger}
			\mb{F}^{*}\right)}{\partial \bs{\psi}}\frac{ \partial \bs{\psi}}{\partial ({}^b\mb{p}_g)} \\[12pt]
		&= \left[
		\frac{\partial}{\partial \bs{\psi}}\left(\mb{J}_{\mb{v} \psi}^{\mathrm{T}}\right)^{\dagger}
		\right]
		\mb{F}^{*}\mb{J}_{\mb{v} \psi}^{\dagger}+\left(\mb{J}_{\mb{v} \psi}^{\mathrm{T}}\right)^{\dagger}\mb{K}_{\mb{\psi}}\mb{J}_{\mb{v} \psi}^{\dagger}
	\end{array}
\end{equation}
where $\left[\frac{\partial}{\partial \bs{\psi}}\left(\mb{J}_{\mb{v}\psi}^{\mathrm{T}}\right)^{\dagger}\right]$ is a $3^{rd}$-order tensor (or a ``3D matrix'').

%% file: content/experiments.tex
\section{Preliminary Experimental Validation}
\label{section:experiments}

To further examine the performance of the proposed aerial continuum manipulator system, a test of arm end-effector displacement under a known external load was conducted. In addition, a benchmark experiment was performed to evaluate the reaction force and torque to the arm's carrier platform while the arm's tip was constrained in the environment. \par

\subsection{Stiffness Test}

Due to the nature of a compliant structure, the continuum arm deflects when external loads applied. It is critical to understand the stiffness of the continuum arm. The external loading test assesses the performance of the arm and can help determine the spectrum of achievable tasks. \par 
Therefore, the displacement of the continuum robot under the action of an external force on its tip has been evaluated to determine the stiffness of the proposed design. The testing fixtures (depicted in Fig.~\ref{fig:stiffness test setup}(a)) was constructed using a group of aluminum profiles with various lengths and certain accessories. The continuum robot's top plate was fastened down to keep the system stationary in space. 
In order to ensure that the external force was applied perpendicularly to the end-disk's norm, a pulley was used to dynamically adjusting the force exerting angle (the blue line in Fig.~\ref{fig:stiffness test setup} indicates the weights hanging cable and its route through pulleys). During outward radial loading experiments, a second pulley could be added to the system to change the direction of hanging weights and make use of gravity. \par
\begin{figure*}[!h]
\setlength\Myfigwd{0.72\columnwidth}
\floatbox[{\capbeside\thisfloatsetup{
capbesideposition={right,center},
capbesidewidth=\dimexpr\linewidth-\Myfigwd-3em\relax}}]{figure}[\FBwidth]
{\caption{Benchmark perching experiment results, reaction force and torque measured by wrist F/T sensor while UR5 end-effector travels in (a) x-direction, and (b) z-direction. Negative force/torque readings indicate that they are applied to the opposite direction and refer to the y-axis on the left of the plot; The red dash line on each plot denotes robot trajectory and refers to the y-axis on the right.}
\label{fig:benchmark results}}
{\includegraphics[width=\Myfigwd]{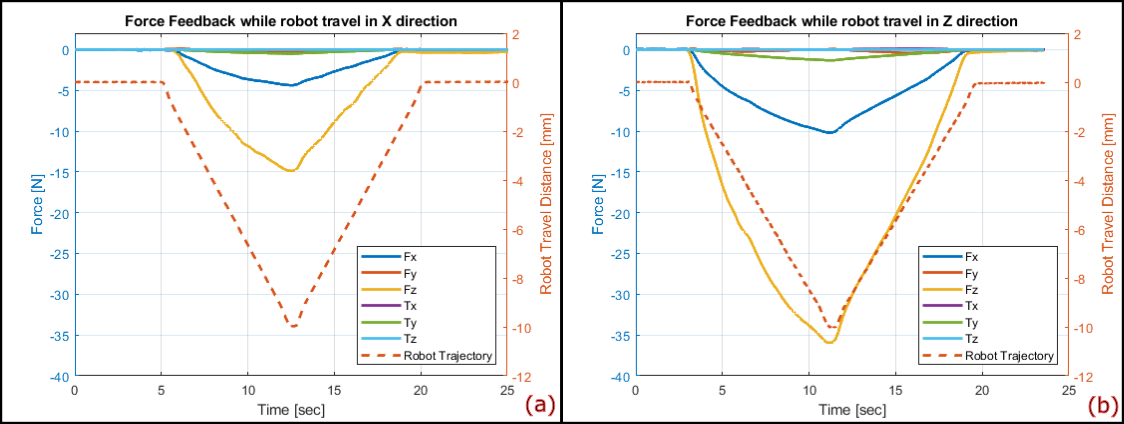}}
\vspace{-6mm}
\end{figure*}

The stiffness tests were performed under two different loading conditions, one was inward radial loading (within the bending plane, load applied with the bending angle), and the other one was outward radial loading (within the bending plane, load applied against the bending angle). An external load of up to approximately 1N was applied perpendicular to the bending direction, with 5 loading cycles and 20 grams increment at each time. The test has been repeated in five different arm configurations to measure the deflection of the aerial continuum arm, and results from inward and outward radial loading are shown in Fig.~\ref{fig:stiffness test setup}(b). As indicated in Fig.~\ref{fig:stiffness results}, the arm straight configuration (denoted in black lines in plots) has the lowest stiffness, and generally as the bending angle increases, the arm stiffens in both loading scenarios.\par

\subsection{Benchmark Perching Experiment}

One advantage of continuum arm is mechanical compliance, which enables a safer robot interaction with the surroundings. While the continuum arm is in motion, it inherently generates a reaction force to its carrier (UAV) that may interfere with the carrier's normal operation. As a result, determining the reaction force caused by arm's motion is useful. \par

Instead of testing the initial arm prototype on an UAV, setting up an in-door test platform is preferable due to its versatility and space efficiency. The test platform can be equipped with various sensors and devices as required by designated experiments.
We use a collaborative robot arm - Universal Robot 5 - as the test platform to simulate the UAV flight motion. UR5 is a 6 degrees-of-freedom industrial robot manipulator with a maximum payload of 5kg. An 6-axis force/torque (F/T) sensor was mounted on UR5's end-effector to sense force transmitted from the tip of continuum arm to the top plate of electronics housing. \par

\begin{figure}
    \vspace{5mm}
    \centering
    \includegraphics[width=0.8\textwidth]{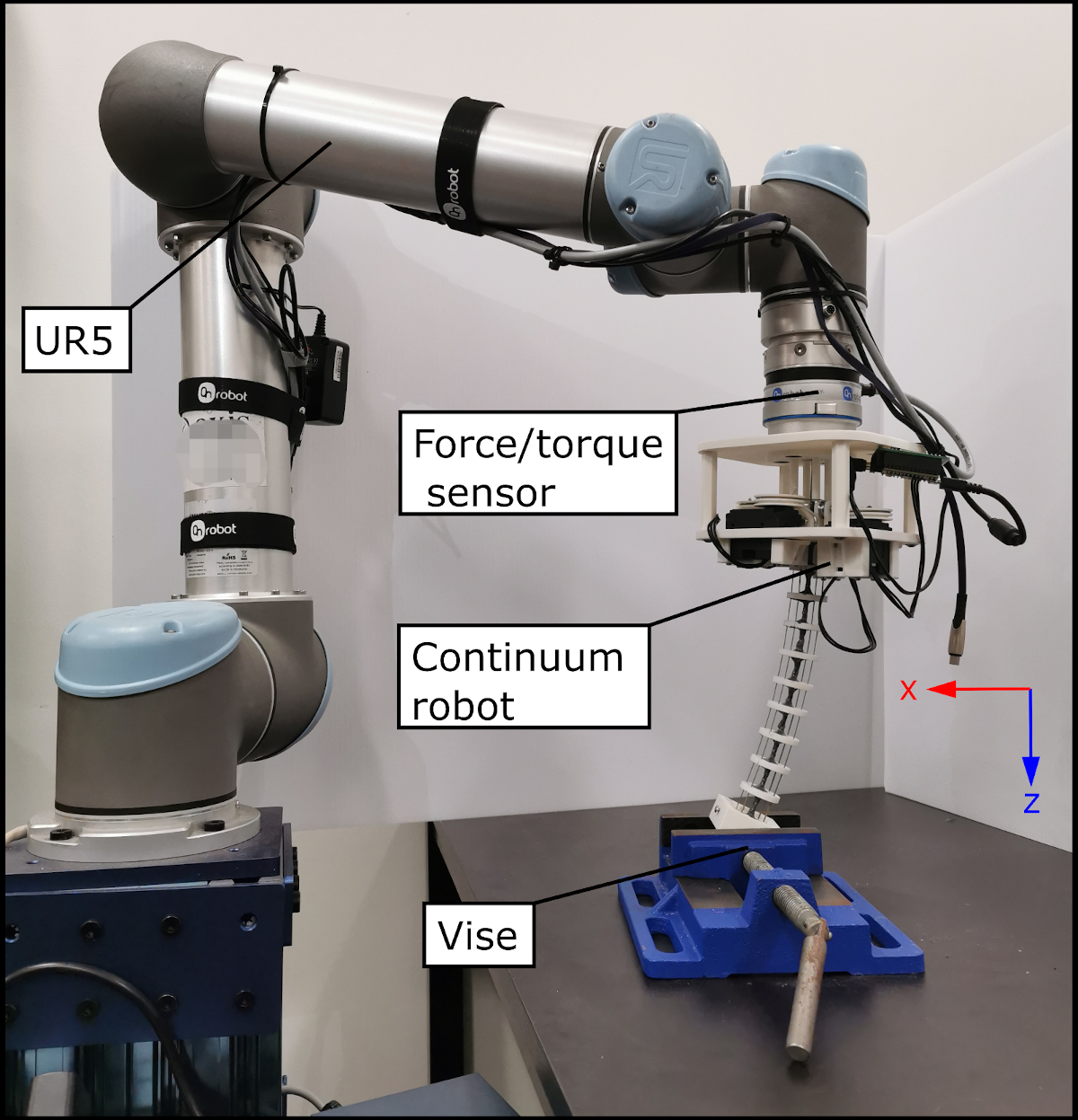}
    \caption{The in-door aerial continuum arm testing platform consists of a UR5 robotic arm, a F/T sensor, and the proposed aerial continuum arm system.}
    \label{fig:benchmark exp setup}
\end{figure}

The benchmark perching experiments evaluate how much force disturbance is exerted to the aerial continuum arm's carrier while the arm tip is constrained in space. As depicted in Fig.~\ref{fig:benchmark exp setup}, the aerial arm system was mounted on the F/T sensor and then on UR5's end-effector. The continuum arm was at a bending angle of 30 degrees, secured at a heavy-duty bench vise on a table. UR5 was instructed to travel 10 mm in its x-direction and then return to its starting position with the same speed. The corresponding force measurements at the F/T sensor were recorded for analysis. The same procedure was repeated with proceeding 10 mm in z-direction and returns. 
Results are shown in Fig.~\ref{fig:benchmark results} (a) and (b) with motion in x and z directions, respectively, with negative force and torque values indicate that they are applied in the opposite direction and refer to the left y-axis of the plot. The red dash line indicates the robot trajectory and refers to the y-axis on the right. As expected, the force exerted to the upper structure of continuum arm system (in this case, the F/T sensor and UR5 end-effector) gradually and smoothly increases and decreases. Results indicate that, when the arm is in contact with environment, the compliance of the continuum arm helps smooth out the force disturbance and avoid a high impact to aerial arm carrier over a short time period. \par

\vspace{-5mm}

%% file: content/discussion.tex
\section{Discussion}

Aerial manipulation is getting an increasingly attention in recent years. The integration of robotic arm and aerial vehicle platform leads to a board range of potential applications. 
The essential goal of aerial manipulation is its accessibility of operating at a high altitude indoor or outdoor workspace that cannot be easily reached by human workers, at the same time to reduce cost and time. Some potential applications include power lines or other infrastructures inspection and maintenance, the transportation of objects, and disaster assistance and rescue. All of these applications required the UAVs to have some physical interactions with the environment and/or human, to operate for a relatively long period of time, to manipulate tools, and to transport goods. 
For the purpose of this proposed aerial continuum arm design and prototype, researchers plan to focus on grasping and perching tasks moving forward, which are the basic functionalities needed by the aforementioned applications. The use of a continuum arm receives much appreciations in that it helps improve robustness of the system and also provides a high tolerance in UAV's pose accuracy. Researchers also aim to add a gripper at the tip of the continuum arm to complete the aerial manipulator for more advanced kinematic and dynamic analysis and experiments. In addition, further investigation on how the continuum arm mechanical design impacts the performance of aerial manipulator is needed.\par

Currently, aerial manipulation activities are tightly limited by aviation administration regulations all over the world. UAVs with certain weight limit must fly in permitted airspaces by certified remote pilots under good weather visibility and within visual line-of-sight. It is researchers responsibility to get familiar with UAV's operation rules and design experiments with rules in mind. \par

%% file: content/conclusion.tex
\section{Conclusion}
\label{section:conclusion}

This paper presents a design of continuum arm system for aerial manipulation that imposes the compliance of a continuum arm structure targeted at perching and grasping applications. Several tests and experiments are carried out to assess the arm's performance when subjected to external force. An in-door experimental setup is built to imitate UAV's motion for the purpose of initial testing and performance measurements. 
However, there is still a lot of work to be done in terms of controlling the arm and the UAV as independent systems and as an integrated system, and then using the system to proceed with more advanced experiments.

%% file: ms.bbl
\begin{thebibliography}{10}

\bibitem{FAA_Forecast}
FAA, accessed December 31, 2019.
\newblock Faa aerospace forecast fiscal years (fy) 2019-2039.

\bibitem{Pounds2011}
Pounds, P. E.~I., Bersak, D.~R., and Dollar, A.~M., 2011.
\newblock ``{Grasping from the air: Hovering capture and load stability}''.
\newblock In 2011 IEEE International Conference on Robotics and Automation,
  IEEE, pp.~2491--2498.

\bibitem{Mellinger2011}
Mellinger, D., Lindsey, Q., Shomin, M., and Kumar, V., 2011.
\newblock ``{Design, modeling, estimation and control for aerial grasping and
  manipulation}''.
\newblock {\em IEEE International Conference on Intelligent Robots and
  Systems}, pp.~2668--2673.

\bibitem{Korpela2013}
Korpela, C., Orsag, M., Pekala, M., and Oh, P., 2013.
\newblock ``{Dynamic stability of a mobile manipulating unmanned aerial
  vehicle}''.
\newblock In 2013 IEEE International Conference on Robotics and Automation,
  IEEE, pp.~4922--4927.

\bibitem{Scholten2013}
Scholten, J.~L., Fumagalli, M., Stramigioli, S., and Carloni, R., 2013.
\newblock ``{Interaction control of an UAV endowed with a manipulator}''.
\newblock {\em Proceedings - IEEE International Conference on Robotics and
  Automation}, pp.~4910--4915.

\bibitem{Korpela2014}
Korpela, C., Orsag, M., and Oh, P., 2014.
\newblock ``{Towards valve turning using a dual-arm aerial manipulator}''.
\newblock In 2014 IEEE/RSJ International Conference on Intelligent Robots and
  Systems, no.~Iros, IEEE, pp.~3411--3416.

\bibitem{Garimella2015}
Garimella, G., and Kobilarov, M., 2015.
\newblock ``{Towards model-predictive control for aerial pick-and-place}''.
\newblock In 2015 IEEE International Conference on Robotics and Automation
  (ICRA), IEEE, pp.~4692--4697.

\bibitem{Anzai2018}
Anzai, T., Zhao, M., Nozawa, S., Shi, F., Okada, K., and Inaba, M., 2018.
\newblock ``{Aerial Grasping Based on Shape Adaptive Transformation by HALO:
  Horizontal Plane Transformable Aerial Robot with Closed-Loop Multilinks
  Structure}''.
\newblock In 2018 IEEE International Conference on Robotics and Automation
  (ICRA), IEEE, pp.~6990--6996.

\bibitem{Nguyen2018}
Nguyen, H.-N., Park, S., Park, J., and Lee, D., 2018.
\newblock ``{A Novel Robotic Platform for Aerial Manipulation Using Quadrotors
  as Rotating Thrust Generators}''.
\newblock {\em IEEE Transactions on Robotics, \textbf{ 34}}(2), apr,
  pp.~353--369.

\bibitem{Ruggiero2018}
Ruggiero, F., Lippiello, V., and Ollero, A., 2018.
\newblock ``{Aerial Manipulation: A Literature Review}''.
\newblock {\em IEEE Robotics and Automation Letters, \textbf{ 3}}(3), jul,
  pp.~1957--1964.

\bibitem{Baraban2020}
Baraban, G., Sheckells, M., Kim, S., and Kobilarov, M., 2020.
\newblock ``{Adaptive Parameter Estimation for Aerial Manipulation}''.
\newblock In 2020 American Control Conference (ACC), Vol.~2020-July, IEEE,
  pp.~614--619.

\bibitem{Kim2020}
Kim, D., and Oh, P.~Y., 2020.
\newblock ``{Testing-and-Evaluation Platform for Haptics-based Aerial
  Manipulation with Drones}''.
\newblock {\em Proceedings of the American Control Conference, \textbf{
  2020-July}}(Figure 2), pp.~1453--1458.

\bibitem{Yuksel2015}
Yuksel, B., Mahboubi, S., Secchi, C., Bulthoff, H.~H., and Franchi, A., 2015.
\newblock ``{Design, identification and experimental testing of a light-weight
  flexible-joint arm for aerial physical interaction}''.
\newblock {\em Proceedings - IEEE International Conference on Robotics and
  Automation, \textbf{ 2015-June}}(June), pp.~870--876.

\bibitem{Suarez2015}
Suarez, A., Heredia, G., and Ollero, A., 2015.
\newblock ``{Lightweight compliant arm for aerial manipulation}''.
\newblock {\em IEEE International Conference on Intelligent Robots and Systems,
  \textbf{ 2015-Decem}}, pp.~1627--1632.

\bibitem{Suarez2018}
Suarez, A., Heredia, G., and Ollero, A., 2018.
\newblock ``{Design of an Anthropomorphic, Compliant, and Lightweight Dual Arm
  for Aerial Manipulation}''.
\newblock {\em IEEE Access, \textbf{ 6}}, pp.~29173--29189.

\bibitem{Samadikhoshkho2020}
Samadikhoshkho, Z., Ghorbani, S., and Janabi-Sharifi, F., 2020.
\newblock ``{Modeling and Control of Aerial Continuum Manipulation Systems: A
  Flying Continuum Robot Paradigm}''.
\newblock {\em IEEE Access, \textbf{ 8}}, pp.~176883--176894.

\bibitem{Fishman2021}
Fishman, J., Ubellacker, S., Hughes, N., and Carlone, L., 2021.
\newblock ``{Dynamic Grasping with a "Soft" Drone: From Theory to Practice}''.
\newblock pp.~4214--4221.

\bibitem{Fishman2021a}
Fishman, J., and Carlone, L., 2021.
\newblock ``{Control and Trajectory Optimization for Soft Aerial
  Manipulation}''.
\newblock In 2021 IEEE Aerospace Conference (50100), Vol.~2021-March, IEEE,
  pp.~1--17.

\bibitem{Hannan2003}
Hannan, M.~W., and Walker, I.~D., 2003.
\newblock ``{Kinematics and the Implementation of an Elephant's Trunk
  Manipulator and Other Continuum Style Robots}''.
\newblock {\em Journal of Robotic Systems, \textbf{ 20}}(2), feb, pp.~45--63.

\bibitem{Liu2019}
Liu, Y., Ge, Z., Yang, S., Walker, I.~D., and Ju, Z., 2019.
\newblock ``{Elephant's Trunk Robot: An Extremely Versatile Under-Actuated
  Continuum Robot Driven by a Single Motor}''.
\newblock {\em Journal of Mechanisms and Robotics, \textbf{ 11}}(5), oct,
  pp.~555--561.

\bibitem{Yeshmukhametov2019}
Yeshmukhametov, A., Koganezawa, K., and Yamamoto, Y., 2019.
\newblock ``A novel discrete wire-driven continuum robot arm with passive
  sliding disc: Design, kinematics and passive tension control''.
\newblock {\em Robotics, \textbf{ 8}}(3).

\bibitem{Zheng2012}
Zheng, T., Branson, D.~T., Kang, R., Cianchetti, M., Guglielmino, E., Follador,
  M., Medrano-Cerda, G.~A., Godage, I.~S., and Caldwell, D.~G., 2012.
\newblock ``{Dynamic continuum arm model for use with underwater robotic
  manipulators inspired by octopus vulgaris}''.
\newblock In 2012 IEEE International Conference on Robotics and Automation,
  IEEE, pp.~5289--5294.

\bibitem{Zheng2013}
Zheng, T., Branson, D.~T., Guglielmino, E., Kang, R., {Medrano Cerda}, G.~A.,
  Cianchetti, M., Follador, M., Godage, I.~S., and Caldwell, D.~G., 2013.
\newblock ``{Model Validation of an Octopus Inspired Continuum Robotic Arm for
  Use in Underwater Environments}''.
\newblock {\em Journal of Mechanisms and Robotics, \textbf{ 5}}(2), may,
  pp.~1--11.

\bibitem{Cianchetti2012}
Cianchetti, M., Follador, M., Mazzolai, B., Dario, P., and Laschi, C., 2012.
\newblock ``{Design and development of a soft robotic octopus arm exploiting
  embodied intelligence}''.
\newblock In 2012 IEEE International Conference on Robotics and Automation,
  IEEE, pp.~5271--5276.

\bibitem{Cianchetti2015}
Cianchetti, M., Calisti, M., Margheri, L., Kuba, M., and Laschi, C., 2015.
\newblock ``{Bioinspired locomotion and grasping in water: the soft eight-arm
  OCTOPUS robot}''.
\newblock {\em Bioinspiration \& Biomimetics, \textbf{ 10}}(3), may, p.~035003.

\bibitem{Gravagne2002}
Gravagne, I., and Walker, I., 2002.
\newblock ``{Manipulability, force, and compliance analysis for planar
  continuum manipulators}''.
\newblock {\em IEEE Transactions on Robotics and Automation, \textbf{ 18}}(3),
  jun, pp.~263--273.

\bibitem{Gravagne2002a}
Gravagne, I., and Walker, I., 2002.
\newblock ``{Uniform regulation of a multi-section continuum manipulator}''.
\newblock In Proceedings 2002 IEEE International Conference on Robotics and
  Automation (Cat. No.02CH37292), Vol.~2, IEEE, pp.~1519--1524.

\bibitem{Simaan2004}
Simaan, N., Taylor, R., and Flint, P., 2004.
\newblock ``{A dexterous system for laryngeal surgery}''.
\newblock In IEEE International Conference on Robotics and Automation, 2004.
  Proceedings. ICRA '04. 2004, Vol.~2004, IEEE, pp.~351--357 Vol.1.

\bibitem{Conrad2015}
Conrad, B., and Zinn, M., 2015.
\newblock ``{Closed loop task space control of an interleaved continuum-rigid
  manipulator}''.
\newblock In 2015 IEEE International Conference on Robotics and Automation
  (ICRA), Vol.~2015-June, IEEE, pp.~1743--1750.

\bibitem{Conrad2017}
Conrad, B.~L., and Zinn, M.~R., 2017.
\newblock ``{Interleaved Continuum-Rigid Manipulation: An Approach to Increase
  the Capability of Minimally Invasive Surgical Systems}''.
\newblock {\em IEEE/ASME Transactions on Mechatronics, \textbf{ 22}}(1), feb,
  pp.~29--40.

\bibitem{simaan2018medical}
Simaan, N., Yasin, R.~M., and Wang, L., 2018.
\newblock ``Medical technologies and challenges of robot-assisted minimally
  invasive intervention and diagnostics''.
\newblock {\em Annual Review of Control, Robotics, and Autonomous Systems,
  \textbf{ 1}}, pp.~465--490.

\bibitem{McMahan2006}
McMahan, W., Chitrakaran, V., Csencsits, M., Dawson, D., Walker, I.~D., Jones,
  B.~A., Pritts, M., Dienno, D., Grissom, M., and Rahn, C.~D., 2006.
\newblock ``{Field trials and testing of the OctArm continuum manipulator}''.
\newblock {\em Proceedings - IEEE International Conference on Robotics and
  Automation, \textbf{ 2006}}(May), pp.~2336--2341.

\bibitem{Rone2018}
Rone, W.~S., Saab, W., and Ben-Tzvi, P., 2018.
\newblock ``{Design, Modeling, and Integration of a Flexible Universal Spatial
  Robotic Tail}''.
\newblock {\em Journal of Mechanisms and Robotics, \textbf{ 10}}(4), aug.

\bibitem{Rone2019}
Rone, W.~S., Saab, W., Kumar, A., and Ben-Tzvi, P., 2019.
\newblock ``{Controller Design, Analysis, and Experimental Validation of a
  Robotic Serpentine Tail to Maneuver and Stabilize a Quadrupedal Robot}''.
\newblock {\em Journal of Dynamic Systems, Measurement, and Control, \textbf{
  141}}(8), aug.

\bibitem{sitler2022modular}
Sitler, J.~L., and Wang, L., 2022.
\newblock ``A modular open-source continuum manipulator for underwater remotely
  operated vehicles''.
\newblock {\em Journal of Mechanisms and Robotics, \textbf{ 14}}(6), p.~061004.

\bibitem{JONES2004}
Jones, B.~A., McMahan, W., and Walker, I., 2004.
\newblock ``Design and analysis of a novel pneumatic manipulator''.
\newblock {\em IFAC Proceedings Volumes, \textbf{ 37}}(14), pp.~687--692.
\newblock 3rd IFAC Symposium on Mechatronic Systems 2004, Sydney, Australia,
  6-8 September, 2004.

\bibitem{Fras2018}
Fras, J., Macias, M., Noh, Y., and Althoefer, K., 2018.
\newblock ``Fluidical bending actuator designed for soft octopus robot
  tentacle''.
\newblock In 2018 IEEE International Conference on Soft Robotics (RoboSoft),
  pp.~253--257.

\bibitem{Xu2021}
Xu, Y., Peyron, Q., Kim, J., and Burgner-Kahrs, J., 2021.
\newblock ``Design of lightweight and extensible tendon-driven continuum robots
  using origami patterns''.
\newblock In 2021 IEEE 4th International Conference on Soft Robotics
  (RoboSoft), pp.~308--314.

\bibitem{Neumann2016}
Neumann, M., and Burgner-Kahrs, J., 2016.
\newblock ``Considerations for follow-the-leader motion of extensible
  tendon-driven continuum robots''.
\newblock In 2016 IEEE International Conference on Robotics and Automation
  (ICRA), pp.~917--923.

\bibitem{Renda_2012}
Renda, F., Cianchetti, M., Giorelli, M., Arienti, A., and Laschi, C., 2012.
\newblock ``A 3d steady-state model of a tendon-driven continuum soft
  manipulator inspired by the octopus arm''.
\newblock {\em Bioinspiration {\&} Biomimetics, \textbf{ 7}}(2), may,
  p.~025006.

\bibitem{McMahan_2005}
McMahan, W., Jones, B., and Walker, I., 2005.
\newblock ``Design and implementation of a multi-section continuum robot:
  Air-octor''.
\newblock In 2005 IEEE/RSJ International Conference on Intelligent Robots and
  Systems, pp.~2578--2585.

\bibitem{simaan2005snake}
Simaan, N., 2005.
\newblock ``Snake-like units using flexible backbones and actuation redundancy
  for enhanced miniaturization''.
\newblock In Proceedings of the 2005 IEEE international conference on robotics
  and automation, IEEE, pp.~3012--3017.

\bibitem{Goldman2014}
Goldman, R.~E., Bajo, A., and Simaan, N., 2014.
\newblock ``Compliant motion control for multisegment continuum robots with
  actuation force sensing''.
\newblock {\em IEEE Transactions on Robotics, \textbf{ 30}}(4), pp.~890--902.

\end{thebibliography}
